\documentclass[times,onecolumn,final,authoryear]{article}

\usepackage{framed,multirow}
\usepackage{authblk}

\usepackage{amssymb}
\usepackage{amsmath}
\usepackage{amsthm}
\usepackage{latexsym}
\usepackage{makecell}
\usepackage{tikz}
\usetikzlibrary{arrows}
\usepackage{subfigure}
\usepackage[utf8]{inputenc}
\usepackage[T1]{fontenc}
\usepackage[linesnumbered,ruled]{algorithm2e}
\usepackage{booktabs}

\usepackage{url}
\usepackage{xcolor}
\definecolor{newcolor}{rgb}{.8,.349,.1}
\theoremstyle{definition}
\newtheorem{definition}{Definition}
\newtheorem{problem}{Problem}
\newtheorem{model}{Model}
\begin{document}

\author[1]{Zhoubo Xu}

\author[1]{Puqing Chen \thanks{c.puqing@gmail.com}}

\author[2]{Romain Raveaux \thanks{romain.raveaux@univ-tours.fr}}

\author[1]{Xin Yang}
\author[1]{Huadong Liu}

\affil[1]{GUET: Guilin University of Electronic Technology, 1 Jinji Road, Guilin, Guangxi, China}
\affil[2]{Université de Tours, Laboratoire d’Informatique Fondamentale et Appliquée de Tours (LIFAT - EA 6300), 64 Avenue Jean Portalis, 37000 Tours, France }

\title{Deep graph matching meets mixed-integer linear programming: Relax at your own risk ?}

%
%
%
%
%

\maketitle              

\begin{abstract}
Graph matching is an important problem that has received widespread attention, especially in the field of computer vision. Recently, state-of-the-art methods seek to incorporate graph matching with deep learning. However, there is no research to explain what role the graph matching algorithm plays in the model. Therefore, we propose an approach integrating a MILP formulation of the graph matching problem. This formulation is solved to optimal and it provides inherent baseline. Meanwhile, similar approaches are derived by releasing the optimal guarantee of the graph matching solver and by introducing a quality level. This quality level controls the quality of the solutions provided by the graph matching solver. In addition, several relaxations of the graph matching problem are put to the test. Our experimental evaluation gives several theoretical insights and guides the direction of deep graph matching methods.
\end{abstract}

\newpage

\section{Introduction}

Matching problem is an important problem, and it appears in many fields such as biology information, chemistry molecule, pattern recognition and computer vision. Especially in the computer vision, matching problem is a crucial issue and frequently occur in stereo matching\cite{luo2016efficient}, target tracking\cite{yilmaz2006object} and pedestrian re-identification\cite{PRD}, to name a few. The matching problem can be decomposed into two parts: a) Extraction of local features from the raw inputs. b) Matching at best the local features. Matching implies solving an assignment problem and solving the conflicting evidence. Matching at best the local features can be modeled as a graph matching problem to consider the relation between local features. 

The graph matching problem is a Quadratic Assignment Problem (QAP) which is known to be $\mathcal{NP}$-hard \cite{GMcomplexity} and it can be solved by combinatorial algorithms \cite{conte2004thirty} and by learning-based methods \cite{caetano2009learning}.  

Deep learning based methods have been applied to graph matching \cite{PCA-GM,DGMC,GMN,SinkhornNet} and are called deep graph matching methods. Deep learning technique can be used to extract visual features and to solve the matching problem in a end-to-end fashion. 

Deep graph matching approaches rely on finding suitable differentiable relaxations of the graph matching problem \cite{PCA-GM,DGMC,GMN,SinkhornNet}. All these approaches have in common that they make sacrifices on the combinatorial side.

On the opposite in \cite{BB-GM}, the authors reduced the sacrifices on the combinatorial side. To reach this goal, they incorporated an efficient combinatorial solver \cite{DBLP:journals/corr/SwobodaRAKS16} which is based on a common dual ascent framework. In \cite{BB-GM}, the authors could integrate the combinatorial solver into the neural network architecture thanks to a recent progress on differentiation of combinatorial solver \cite{DBLP:journals/corr/abs-1912-02175}. However, this combinatorial solver \cite{DBLP:journals/corr/SwobodaRAKS16} is still a heuristic method and cannot guarantee that the optimal solution of the graph matching problem is found.

From the above statements, important questions appear: Do exact methods matter in the context of deep graph matching? Can they improve the matching accuracy with respect to the ground truth? Are heuristics sufficient because the learned features makes the conflicting evidence disappear? In the learned feature space, is the matching problem so simple that it can be solved efficiently by a trivial heuristic? However, there are not any theoretical explanations or practical proofs. So, we hypothesize that the optimal solution of the graph matching problem is the best solution to the matching task as long as matching scores reflect the user need or the task objective. Such a case appears in a end-to-end framework trained in a supervised manner.
We hypothesize that low-quality solutions provided by a graph matching solver can prevent or impact negatively the convergence of the training algorithm. These two hypotheses are put to the test in this paper. 

In this work, we present a novel end-to-end architecture for semantic key point matching that does not make any concessions on the combinatorial side. For the first time, an exact graph matching solver is embedded into a deep learning architecture. Especially, we focus on a Mixed-Integer Linear Programming (MILP) that models, in a compact manner, the graph matching problem\cite{Lerouge2017NewBL}. Optimal solutions are obtained by a MILP solver. It takes advantage of the mathematical structure of the graph matching problem. In addition, the MILP solver is tuned to compute suboptimal solutions according to a desired quality level. With our paper, we aim to answer if a heavily optimized exact graph matching solver can increase matching performances in the context of deep graph matching. In addition, we analyze the impact of suboptimal solutions of different qualities. Finally, the impact of different relaxations of the original graph marching problem is also studied.

The contributions presented in this paper can be summarized as follows:
\begin{itemize}
    \item We present an end-to-end trainable architecture that incorporates an exact graph matching algorithm in a mathematically sound fashion, which named DIP-GM. The algorithm achieves state-of-the-art results in the standard key points matching data sets Pascal VOC\cite{bourdev2009poselets,everingham2010pascal} (with Berkeley annotations) and SPair-71K\cite{min2019spair}. We argue that the exact graph matching algorithm is a more inherent choice as the baseline of the deep graph matching method compared to the heuristic algorithm.
    
    \item In order to explore the impact of different quality solutions of the MILP solver on the model, we derive from the MILP solver a quality-aware heuristic and test on the Pascal VOC (with Berkeley annotations\cite{bourdev2009poselets,everingham2010pascal}) and the SPair-71K\cite{min2019spair}. The experiment results proof that the exact method is important on the deep graph methods. However, approximate in a certain range will not bring negative effects to the model, and can even increase the matching accuracy.
    
    \item We propose a framework to compare different relaxations of the original graph matching solvers. Two types of relaxations are considered. a) The graph matching problem is reduced from a quadractic assignement problem to linear assignement problem. b) Topological constraints of the graph matching problem are relaxed to match vertices and edges independently. The impact of the relaxations is studied to measure the risk on the matching task. 
    
    \item In the same framework, we also implemented the continuous solver named Sinkhorn \cite{SinkhornAlgo} for a continuous relaxation of the graph matching problem. Through this way, we can compare combinatorial and continuous solvers when they are integrated into a deep learning architecture. 
\end{itemize}

Next, we introduce the rest of the paper.  Section \ref{sec:problems}, we give the definitions of the problems to be studied in this paper. Section \ref{sec:stateofheart}, the state of the art on graph matching is presented. Section \ref{sec:proposal}, our proposal is described. A neural network architecture that integrates a MILP solver is presented. Section 5 shows several experiment details and results. Finally, Section 6 presents our conclusions and open research directions inspired by our initial findings.

\section{Problems}
\label{sec:problems}
In this section, we define the problems to be studied. First, the graph matching problem is described. Second, the supervised learning problem is presented in the context of learning a matching function. Finally, supervised learning can be achieved by gradient descent-like algorithms. In such cases, a new problem appears called Informative gradient problem.

\subsection{Graph matching problem}
An attributed graph is considered as a pair 
$(V ,E)$ such that: $G = (V ,E)$. $G$ belongs to the graph space denoted $\mathbb{G}$. $V$ is a set of vertices. $E$ is a set of edges such as $E \subseteq V \times V$.
The objective of graph matching is to find correspondences between two attributed graphs $G_1=(V_1,E_1)$ and $G_2=(V_2,E_2)$. A solution of graph matching is defined as a subset of possible correspondences $\mathcal{V} \subseteq V_1 \times V_2$, represented by a binary assignment matrix $ \textbf{V} \in \{0,1 \}^{|V_1| \times |V_2|}$. If $u_i \in V_1$ matches $u_k \in V_2$, then $\textbf{V}_{i,k}=1$, and $\textbf{V}_{i,k}=0$ otherwise. We
denote by $\textbf{v} \in \{0,1 \}^{|V_1| . |V_2|}$, a column-wise vectorized replica of $\textbf{V}$. With this notation, graph matching problems can be expressed as finding the assignment vector $\hat{\textbf{v}}$ that maximizes a score function $S(G_1, G_2,  \textbf{v})$ as follows:
\begin{problem}{Graph Matching Problem}
\label{prob:GM}
\begin{subequations}
  \begin{align}
   \hat{\textbf{v}} =& \underset{\textbf{v}} {\mathrm{argmax}} \quad S(G_1, G_2, \textbf{v})\\
    \text{subject to}\quad & \textbf{v} \in \{0, 1\}^{^{|V_1| . |V_2|}}\\
    \label{eq:subgmgmc}
    &\sum_{u_i \in V_1} \textbf{v}_{i,k} \leq 1 \quad \forall u_k \in V_2\\
    \label{eq:subgmgmd}
     &\sum_{u_k \in V_2} \textbf{v}_{i,k} \leq 1 \quad \forall u_i \in V_1
  \end{align}
\end{subequations}
\end{problem}

where equations \eqref{eq:subgmgmc},\eqref{eq:subgmgmd} induces the matching constraints, thus making $\mathbf{v}$ an assignment vector.

The function $S(G_1, G_2, \textbf{v})$ measures the similarity of graph attributes, and is typically decomposed into a first order similarity function $s(u_i, u_k)$ for a node pair $u_{i} \in V_1$ and $u_k \in V_2$, and a second-order similarity function $s(z_{ij}, z_{kl})$ for an edge pair $z_{ij} \in E_1$ and $z_{kl} \in E_2$. Thus, the objective function of graph matching is defined as: 
  \begin{equation}
  \label{eq:matchingfunction}
  \begin{aligned}
  S(G_1,G_2,\textbf{v}) =&	\sum_{u_i \in V_1}\sum_{u_k \in V_2} s(u_i, u_k)  \cdot \textbf{v}_{i,k} \\ 
  &+ \sum_{z_{ij} \in E_1}\sum_{z_{kl} \in E_2} s(z_{ij}, z_{kl}) \cdot  \textbf{v}_{i,k} \cdot \textbf{v}_{j,l}
    \end{aligned}
\end{equation}
 
In essence, the score accumulates all the similarity values that are relevant to the assignment. $S$ is a quadratic function according to the variable $\textbf{v}$. The problem has been proven to be $\mathcal{NP}$-hard by \cite{GMcomplexity}.

Similarities $\{$ $s(u_i, u_k)$ $\vert$ $u_{i} \in V_1$ and $u_k \in V_2$ $\}$ can be stored in a vector $\mathbf{sv}  \in \mathbb{R}^{|V_1| \cdot |V_2|}$. Similarities
 $\{$ $s(z_{ij}, z_{kl})$ $\vert$ $z_{ij} \in E_1$ and $z_{kl} \in E_2$ $\}$ can be stored in a vector $\mathbf{se}  \in \mathbb{R}^{|E_1| \cdot |E_2|}$.

\begin{equation}
  \label{eq:matchingfunctionquadvecto}
  \begin{aligned}
  S(\textbf{v},\mathbf{sv},\mathbf{se}) =& \mathbf{sv}^{T} \cdot \mathbf{v} + 
  \sum_{z_{ij} \in E_1}\sum_{z_{kl} \in E_2} \textbf{v}_{i,k} \cdot \textbf{v}_{j,l} \cdot \mathbf{se}_{ij,kl}
    \end{aligned}
\end{equation}

\subsection{Learning graph matching problem}
\label{sec:subsec:learninggm}
The learning graph matching problem can be framed an error minimization problem. The error is measured by a loss function ($L$). For a pair of graphs to be matched, $L$ is a measure of the differences between a predicted matching ($\hat{\mathbf{v}}$) and a ground-truth's matching $\textbf{v}^{gt}$. The learning graph matching problem can be stated as minimizing the average loss function on a data set of graph pairs. The training set is then defined as $TrS=\{ ((G_1,G_2)_k,\textbf{v}^{gt}_k) \}_{k=1}^M$. $(G_1,G_2)$ is a graph pair. Generally speaking, the learning problem can be stated as follows:
\begin{problem}{Learning graph matching problem}
\begin{equation}
\hat{\theta}=arg \min_{\theta \in \Theta}  \sum_{ ((G_1,G_2),\textbf{v}^{gt}) \in TrS} \quad  L(\textbf{v}^{gt},\Phi(\theta;G_1,G_2))  
\end{equation}
Where $\Phi: \mathbb{G} \times \mathbb{G} \times \Theta \to \Gamma$ is a mapping function. $\Phi$ takes as an input a graph pair $(G_1 \in \mathbb{G}, G_2 \in \mathbb{G})$ and the trainable parameters $\theta$.  $\theta$ are the variables of the learning problem. They are trainable parameters (invariants) over the data set $TrS$. $\Phi$ outputs a matching ($\hat{\mathbf{v}} \in \Gamma$). Such that $\Gamma$ is the set of all possible graph matchings between $G_1$ and $G_2$.
$L$ is the loss function and a possible choice is the Hamming distance $L=\Vert \textbf{v}^{gt} - \hat{\textbf{v}} \Vert_1 $.  
\label{prob:gmlearning}
\end{problem}
Inside this problem, two learning problems arise: 
\begin{itemize}
\item Node/edge attributes can be learned. This procedure is assimilated to a feature extraction step. In a more limited way, learning algorithms can only be applied to learn node/edge similarity functions.
\item Finally, the graph matching algorithm itself can be improved or replaced by a learnable function.
\end{itemize}

\subsection{Learning by gradient descent}
The learning graph matching problem (Problem \ref{prob:gmlearning}) can be solved in a end-to-end fashion by a neural network architecture trained by gradient descent. In such a case, let us define the specific problem that arises when learning involved a combinatorial problem. 

Let $NN$ be a function representing the neural network. $G_1 \in \mathbb{G}$ and $G_2 \in \mathbb{G}$ are the graphs and $\theta \in \Theta$ are the parameters of the neural network.
\begin{definition}{Neural Network $NN$}
\begin{equation}
NN : \mathbb{G} \times \mathbb{G} \times \Theta  \to [\mathbb{R}^N,\mathbb{R}^n]
\end{equation}
\begin{equation}
NN(G_1,G_2,\theta)  \mapsto [\mathbf{sv} \in \mathbb{R}^{|V_1| \cdot |V_2|} , \mathbf{se} \in \mathbb{R}^{|E_1| \cdot |E_2|}]
\end{equation}
\end{definition}

According to the problem \ref{prob:GM} and the equation \eqref{eq:matchingfunctionquadvecto}, the graph matching solver receives continuous inputs $\mathbf{sv}$ and $\mathbf{se}$ and return discrete output $\hat{\textbf{v}}$ from some finite set $\Gamma$.
\begin{equation}
GMS :  [\mathbf{sv}, \mathbf{se}]  \mapsto \mathbf{ \hat{v} }
\end{equation}

$L$ is the loss function of the neural network architecture. $L$ can be defined as a composition of functions.
\begin{definition}{Loss function L}
\begin{equation}
L : \Gamma \times \Gamma
\end{equation}
\begin{equation}
(L  \circ GMS \circ NN)(G_1,G_2,\mathbf{v}^{gt})=L(\mathbf{v}^{gt},GMS(NN(G_1,G_2,\theta)))
\end{equation}
\end{definition}
Where $\mathbf{v}^{gt} \in \Gamma$ is a ground-truth matching solution given by an oracle.

The computation of $L$ is possible thanks to the \textit{forward algorithm}. It means that the input data is fed in the forward direction through the network. The forward pass is denoted by $GMS(NN(G_1,G_2,\theta))$.

This learning problem is to find $\theta$ such that $L$ is minimum for a given data set composed of samples $(G_1,G_2,\mathbf{v}^{gt})$. 
\begin{definition}{Learning graph matching problem with a neural network}
\begin{equation}
\theta^* = arg \min_{\theta} \sum_{ ((G_1,G_2),\textbf{v}^{gt}) \in TrS} \quad
 L(\mathbf{v}^{gt},GMS(NN(G_1,G_2,\theta)))
\end{equation}
\end{definition}
The learning problem is solved by finding where the derivative equal to zero.

\begin{definition}{Learning problem formulated as a gradient-based problem}
\begin{equation}
\text{Find } \theta^* \text{ such that } \sum_{ ((G_1,G_2),\textbf{v}^{gt}) \in TrS} \quad \frac{\partial L(\mathbf{v}^{gt},GMS(NN(G_1,G_2,\theta^*)))}{\partial \theta^*} =0
\end{equation}
\end{definition}
Finding where the derivative equal to zero can be achieved by gradient descent methods. Gradient descent methods are iterative methods and they can be formulated as follows: 
$$\theta^{(t+1)}= \theta^{(t)} - \; lr. \, \sum_{ ((G_1,G_2),\textbf{v}^{gt}) \in TrS} \quad
\frac{\partial L(\mathbf{v}^{gt},GMS(NN(G_1,G_2,\theta)))}{\partial \theta} $$
The learning rate ($lr$) is a parameter that controls the speed of the descent.

The gradient of the loss $L$ with respect to the parameters $\theta$ is given by the chain rule of derivatives as follows:
\begin{definition}{Gradient of the loss}
\begin{equation}
\label{eq:gradlossgeneral}
\frac{\partial L(\theta)}{\partial \theta} =  \frac{\partial L(\mathbf{ {v} })}{\partial \mathbf{ {v} } } . \frac{\partial GMS( \mathbf{sv}, \mathbf{se})}{\partial  [\mathbf{sv}, \mathbf{se}] } . \frac{\partial NN(\theta)}{\partial \theta }
\end{equation}
\end{definition}
This chain of derivatives is also called \textit{back-propagation}.

The crucial issue is to compute an informative gradient from $ \frac{\partial GMS( \mathbf{sv}, \mathbf{se})}{\partial  [\mathbf{sv}, \mathbf{se}] }$.

\begin{problem}{Informative gradient problem}
\label{prob:gradprob}
$$\frac{\partial GMS( \mathbf{sv}, \mathbf{se})}{\partial  [\mathbf{sv}, \mathbf{se}] } \sim \textbf{0}$$

The $GMS$ contains an $argmax$ operator and discrete variables that prevent the $GMS$ to be sensitive to small perturbations. The fundamental problem with differentiating through combinatorial solvers like $GMS$ is not the lack of differentiability; the gradient of the loss $L$ exists almost everywhere. However, this gradient of the loss $L$ is a constant zero and is unhelpful for optimization by gradient-descent methods. 
\end{problem}

\section{State of the art}
\label{sec:stateofheart}
First, we present the solution methods for the graph matching problem (Problem \ref{prob:GM}) and finally, approaches to solve the learning graph matching problem (Problem \ref{prob:gmlearning}) are detailed.

\subsection{Graph matching solvers}
The graph matching problem has been proven to be $\mathcal{NP}$-hard. So, unless  $\mathcal{P} = \mathcal{NP}$, solving the problem to optimality cannot be done in polynomial time of the size of the input graphs. Consequently, the runtime complexity of exact methods is not polynomial but exponential with respect to the number of vertices of the graphs. On the other hand, heuristics are used when the demand for low computational time dominates the need to obtain optimality guarantees.

 Many solver paradigms were put to the test for the graph matching problem. These include relaxations based on Lagrangean decompositions \cite{DBLP:journals/corr/SwobodaRAKS16,messagepassingdualdecomposition}, convex/concave quadratic \cite{gnccp} (GNCCP) and semi-definite programming \cite{pbmatching} , which can be used either directly to obtain approximate solutions or just to provide lower bounds. To tighten these bounds several cutting plane methods were proposed \cite{Bazaraa1982}. On the other side, various primal heuristics, both (i) deterministic, such as graduated assignment methods \cite{graduatedassignment}, fixed point iterations \cite{IPFP} (IPFP), spectral technique and its derivatives \cite{SMACGM,spectralmatching} and (ii) non-deterministic (stochastic), like random walk \cite{reweightedgm} were proposed to provide approximate solutions to the problem. 
 
 Especially, two effective graph matching solvers can be further described : 
 \begin{enumerate}
     \item An effective heuristic is presented in \cite{DBLP:journals/corr/SwobodaRAKS16}. It is based Lagrangean decompositions and on gradient ascent method.  On the high level, the idea of the graph matching solver is to decompose the original problem into several ''easier" subproblems, for which an efficient global minimum (or a good lower bound) can be computed. Combining the lower bounds for individual subproblems will then provide a lower bound for the original problem. Lagrange multipliers connecting these subproblems are updated with the sub-gradient method. Gradient descent is often performed by blocks of coordinates (not all the vector at once). The method  proposed by \cite{DBLP:journals/corr/SwobodaRAKS16} is included inside a method of our benchmark. 
     \item An efficient exact method is described in \cite{Lerouge2017NewBL}. The graph matching problem is modelled as an mixed-integer linear program with a
    linear objective function of the matching variables and linear constraints. The mixed-integer linear program is solved by a black box solver called CPLEX \cite{cplex2009v12}. The method proposed by \cite{Lerouge2017NewBL} will be involved in our proposed method.
 \end{enumerate}
 
\subsection{Learning graph matching methods}
Studies \cite{caetano2009learning,cho2013learning} have revealed that hand-crafted matching similarities, which are typically used in graph matching, are insufficient for capturing the inherent structure that underlies the problem at hand. Consequently, better optimization of the graph matching problem does not guarantee a better correspondence relationship with respect to real-world matching. To tackle this issue, machine learning and deep learning techniques have been used. Matching similarities must reflect the user's need; thus, they can be learned to fit a specific goal.

Caetano, et al \cite{caetano2009learning} first treated graph matching problems as a supervised learning problem (Problem \ref{prob:gmlearning}). The learning problem is finding a parameterized objective function that minimizes the normalized hamming loss between predict matching matrix and ground-truth matrix. A combinatorial solver based column-generation was embedded into the method. Caetano, et al works aimed at learning shared parameters among all the edge features and node features. To overcome this limitation, the discriminative weight formulation was introduced by \cite{cho2013learning}. This formulation can assign a vector of parameters to each node and edge of the graph to be matched. The previous works \cite{caetano2009learning,cho2013learning} rely on hand-crafted features and shalow models.

Deep learning has achieved a remarkable performance breakthrough in several fields, most notably in speech recognition, natural language processing, and computer vision. Consequently, deep learning method have been applied to solve the graph matching problem. 

As mentioned in Section \ref{sec:subsec:learninggm}, the learning graph matching problem (Problem \ref{prob:gmlearning}) can be divided into two parts: a) extracting local features from the original input; b)solving the assignment problem. 
The first part gathers methods that focus on the improvement of the feature extraction stage thanks to effective graph neural layers, convolution layers, attention mechanisms or different losses. This group of methods is very generic and is not dedicated to graph matching so we will not focus on it. We will concentrate on the second group of methods that is dedicated to graph matching. The second group of methods is concerned by the integration of the graph matching solver into the neural network architecture. The objective is to better solve the conflicting assignments. The literature on this topic can then be split into two parts whether the approaches rely on a continuous relaxation of the graph matching problem or not.

\paragraph{Relaxation-based methods}
Andrei Zanfir and Cristian Sminchisescu \cite{GMN} proposed a deep learning end-to-end model(GMN) for graph matching. They relax the quadratic assignment problem to the continuous domain and the matching constraints are replaced by a ball constraint forcing the solution to be at the surface a ball of radius one.
\begin{problem}{Relaxed Graph Matching}
\label{prob:relaxgm}
\begin{subequations}
  \begin{align}
   \hat{\textbf{v}} =& \underset{\textbf{v}} {\mathrm{argmax}} \quad S(G_1, G_2, \textbf{v})\\
    \text{subject to}\quad & \textbf{v} \in [0, 1]^{^{|V_1| . |V_2|}}\\
    &|| \textbf{v} ||_2=1
  \end{align}
\end{subequations}
\end{problem}
Problem \ref{prob:relaxgm} is solved by the power iteration method. This model is the first baseline of deep graph matching and it sheds some light for graph matching research. In the same direction, in \cite{SinkhornNet,PCA-GM}, the graph matching is further relaxed into a Linear Sum Assignment Problem (LSAP).

\begin{problem}{Linear Sum Assignment Problem (LSAP)}
\label{prob:lsap}
\begin{subequations}
  \begin{align}
   \hat{\textbf{v}} =& \underset{\textbf{v}} {\mathrm{argmax}} \quad \sum_{u_i \in V_1}\sum_{u_k \in V_2} s(u_i, u_k)  \cdot \textbf{v}_{i,k} \\ 
    \text{subject to}\quad & \textbf{v} \in \{0, 1\}^{^{|V_1| . |V_2|}}\\
    &\sum_{u_i \in V_1} \textbf{v}_{i,k} \leq 1 \quad \forall u_k \in V_2\\
     &\sum_{u_k \in V_2} \textbf{v}_{i,k} \leq 1 \quad \forall u_i \in V_1
  \end{align}
\end{subequations}
\end{problem}

This problem (Problem \ref{prob:lsap}) is a discrete problem with an $\textit{argmax}$ operator so it is proned to the informative gradient problem (Problem \ref{prob:gradprob}). 

Fortunately, the optimal solution of the continuous relaxation of the LSAP gives the optimal solution the LSAP. It can be seen as a consequence of the Birkhoff theorem. It ensures that doubly-stochastic matrices are the convex envelope of permutation matrices.

\begin{problem}{Relaxed Linear Sum Assignment Problem}
\label{prob:relaxedlsap}
\begin{subequations}
  \begin{align}
   \hat{\textbf{v}} =& \underset{\textbf{v}} {\mathrm{argmax}} \quad \sum_{u_i \in V_1}\sum_{u_k \in V_2} s(u_i, u_k)  \cdot \textbf{v}_{i,k} \\ 
    \text{subject to}\quad & \textbf{v} \in [0, 1]^{^{|V_1| . |V_2|}}\\
    &\sum_{u_i \in V_1} \textbf{v}_{i,k} \leq 1 \quad \forall u_k \in V_2\\
     &\sum_{u_k \in V_2} \textbf{v}_{i,k} \leq 1 \quad \forall u_i \in V_1
  \end{align}
\end{subequations}
\end{problem}

The Relaxed LSAP is a special case of the Kantorovich problem (also called optimal transport problem) and it can be solved efficiently thanks to the Sinkhorn algorithm \cite{SinkhornAlgo}. Such an approach is called Sinkhorn Net \cite{PCA-GM} in the literature.

Matthias Fey, et.al\cite{DGMC} extended the Relaxed LSAP (Problem \ref{prob:relaxedlsap}) by adding neighborhood consensus constraints. Such a neighborhood consensus avoids adjacent vertices in the source graph from being mapped to different ''regions" in the target graph. 

Formally, a neighborhood consensus is reached if for all node pairs $(u_i, u_k) \in V_1 \times V_2$ with $\textbf{v}_{i,k}=1$, it holds that for every node $u_j \in \mathcal{N}(u_i)$ there exists a node $u_l \in \mathcal{N}(u_k)$ such that $\textbf{v}_{j,l}=1$. $\mathcal{N}$ is a neighborhood function that returns the adjacent nodes from a root node.

All the prior methods operate with continuous relaxations of the graph matching problem to avoid the informative gradient problem caused by the discrete graph matching problem. 

The consequence is a sacrifice on the combinatorial aspect of the original graph matching problem. This sacrifice leads to a loss of the relations between local features.

\paragraph{Combinatorial-based methods}
To not sacrifice on the combinatorial side, Michal Rol'inek et.al\cite{BB-GM} integrated a graph matching solver based Lagrangean decompositions \cite{DBLP:journals/corr/SwobodaRAKS16}. The integration of this graph matching solver is possible thanks to the work of \cite{DBLP:journals/corr/abs-1912-02175} that seamlessly embeds blackbox implementations of a wide range of combinatorial algorithms into deep networks in a mathematically sound fashion.

The idea developed in \cite{BB-GM} is promising however the graph matching solver is an heuristic method and so the optimal solution is not guarantee to be found and neither the impact of the suboptimal solutions is discussed.

\paragraph{Positioning our proposal}
Effectiveness of deep graph matching methods can be impacted by the quality of the solutions provided by the graph matching solver. The quality of the solutions depends on the effectiveness of the graph matching solver and the type of relaxations of the original graph matching problem.

Judging from all the above work, a very important question is how the quality of the solution provided by the graph matching solver will affect the model training. In view of this, we explore the impact of graph matching solvers on deep graph matching. Our architecture will stand on the shoulder of the architecture developed by \cite{BB-GM}. But, we will replace the heuristic solver \cite{DBLP:journals/corr/SwobodaRAKS16} by an exact method to obtain optimal solutions. In addition, we will gradually degrade the matching solutions thanks to a quality-aware heuristic to clearly measure the evolution of the model effectiveness with respect to solution qualities. Finally, in a fully uniform manner, we will measure the impact of different relaxations of the original graph matching problem. The relaxations to be studied are : 
\begin{enumerate}
    \item The quadratic matching problem is relaxed to a linear assignment problem where only vertices are matched.
    \item The quadratic matching problem is relaxed to a linear assignment problem where  vertex sets and edge sets are matched independently.
    \item Matching variables are relaxed from discrete to continuous domain.
\end{enumerate}

\section{Proposal : a mixed-integer linear program for deep graph matching}
\label{sec:proposal}
In this section, our proposal is described. First, we show a global picture of our architecture with main components. Second, we present the MILP for the graph matching problem. Third, we explain how the MILP can be solved by exact methods. In addition, we show how exact methods can be modified to build an effective heuristic with theoretical guarantees on the quality of solutions. Fourth, when dealing with combinatorial solvers, the informative gradient problem  (Problem \ref{prob:gradprob}) appears. We solve this problem by continuous interpolation of the loss function. We explain how to compute such an interpolation in the case of the graph matching problem. Finally, the integration of the graph matching solver into the neural network is completed by two algorithms that describe, respectively, the forward pass and the back-propagation pass. These two functions are called when a neural network architecture is trained by gradient-descent methods.

\subsection{Architecture design}
Our end-to-end trainable architecture for keypoint matching consists of three stages. We call it ''DIP-GM". The global architecture is displayed in Figure \ref{fig:deepgmarchitecture} and detailed as follows:

\begin{figure}[htbp]
    \centering
    \includegraphics[scale=0.25]{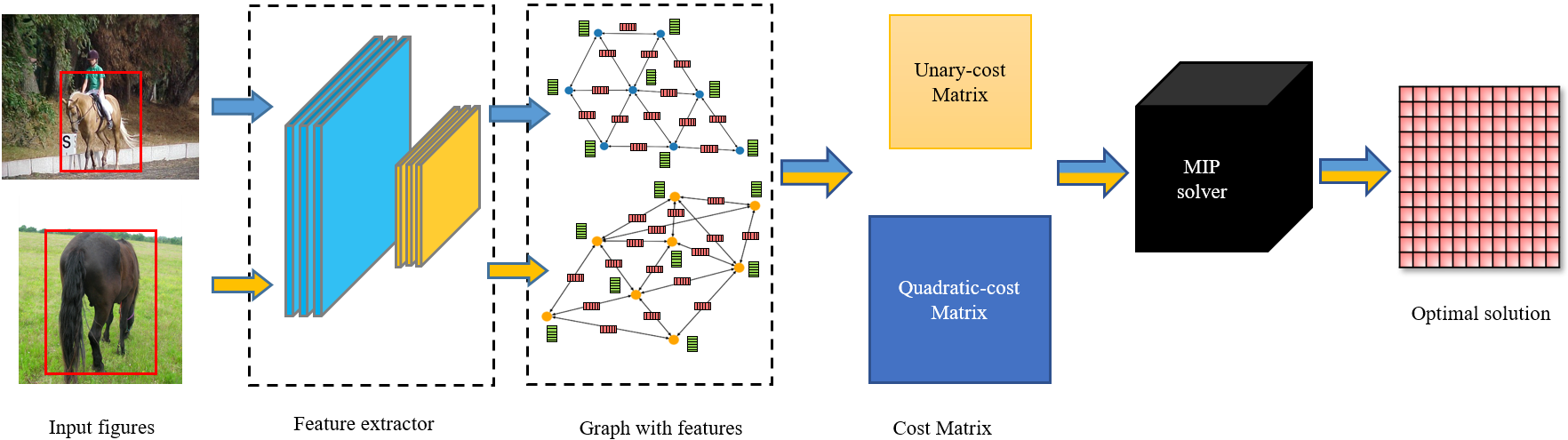}
    \caption{Deep graph matching end-to-end pipeline model with MILP solver.The graphs for matching are construct by Delaunay triangulation\cite{lee1980two}. Features of graph extracted from  VGG16 pretrained with ImageNet \cite{deng2009imagenet} and Spline CNN\cite{fey2018splinecnn}. Similarity matrices will be fed into the $GMS$ (Model \ref{model:F2}) and output an optimal integer solution.}
    \label{fig:deepgmarchitecture}
\end{figure}

\begin{enumerate}
    \item Extraction of visual features A standard CNN architecture extracts a feature vector for each of the keypoints in the image. Additionally, a global feature vector is extracted.
    \item Geometry-aware feature refinement Keypoints are converted to a graph structure with spatial information. Then a graph neural network architecture is
applied. 
   \item At the end of this stage, vertex-to-vertex similarities ($\mathbf{sv}$) and edge-to-edge similarities ($\mathbf{se}$) are computed using the graph features and the global features. This determines a graph matching instance that is passed to the solver.
\end{enumerate}
Our architecture is based on the architecture developed by \cite{BB-GM} that is one of the best method in the state of the art. However from \cite{BB-GM}, we replace the heuristic solver \cite{DBLP:journals/corr/SwobodaRAKS16} by an exact method. We also propose to derive a quality-aware heuristic from the exact method to obtain an heuristic with guarantees on the closeness to the optimal solution. 

\subsection{Mixed-integer linear program for graph matching}
In \cite{Lerouge2017NewBL}, a MILP is proposed to model the graph edit distance problem that is another graph matching problem. However, graph matching and graph edit distance problems have been unified in \cite{DBLP:journals/prl/Raveaux21}. The graph edit distance problem is a cost minimization problem while the graph matching problem (Problem \ref{prob:GM}) is similarity maximization problem.
Consequently as proven in \cite{DBLP:journals/prl/Raveaux21}, the MILP proposed in \cite{Lerouge2017NewBL} can model the graph matching problem by changing the MILP from a minimization to a maximization problem. The MILP has been showed to be efficient because it is compact in terms of variables and constraints. 
Our MILP is described in four steps. First the input data of an instance of the MILP are presented. Second, the variables are listed. Third, the objective function is stated. Finally, constraints are explained.

\paragraph{Data}
 Node similarities $s(u_i, u_k) \; \forall u_i \in V_1, u_k \in V_2$ can be stored in a vector $\mathbf{sv} \in \mathbb{R}^{|V_1|. |V_2|}$. Edge similarities $s(z_{ij}, z_{kl}) \; \forall z_{ij} \in E_1, z_{kl} \in E_2$ can be stored in a vector $\mathbf{se} \in \mathbb{R}^{|E_1|. |E_2|}$.


\paragraph{Variables}
 For nodes and edges matching, a set of binary variables is defined in Table \ref{tab:ILPvar}.
 If $u_i \in V_1$ matches $u_k \in V_2$, then $\textbf{v}_{i,k}=1$, and $\textbf{v}_{i,k}=0$ otherwise.
 If an edge $z_{ij} \in E_1$ is matched to an edge $z_{kl} \in E_2$, then $\textbf{e}_{ij,kl}=1$, and $\textbf{e}_{ij,kl}=0$ otherwise.

\begin{table}[htbp]
    \centering
    \caption{Definition of binary variables of the MILP.}
    \begin{tabular}{|c||c|c|}
    \hline
         Name & Card & Role  \\\hline\hline
         $\mathbf{v}_{i,k}$& $\forall (u_i,u_k) \in V_1 \times V_2$ & =1 if $u_i$ is matched with $u_k$ and 0 otherwise\\\hline
         $ \mathbf{e}_{ij,kl}$& $\forall (z_{ij},z_{kl}) \in E_1 \times E_2$ & =1 if $z_{ij}$ is matched with $z_{kl}$ and 0 otherwise\\\hline
    \end{tabular}
    \label{tab:ILPvar}
\end{table}

 \paragraph{Objective function}
 It is possible to linearize the function $S$ (Equation \eqref{eq:matchingfunction}) at the cost of adding a new variable $\textbf{e} \in \{0,1 \}^{|E_1| . |E_2|}$ as the indicator vector of matched edges.
\begin{equation}
  \label{eq:matchingfunctionlinear}
  \begin{aligned}
  SL(\textbf{v},\textbf{e},\mathbf{sv},\mathbf{se}) =& \mathbf{sv}^{T}.\mathbf{v} +  \mathbf{se}^{T}.\mathbf{e}
    \end{aligned}
\end{equation}
The function $SL$ is a linear function with respect to $\mathbf{v}$ and $\mathbf{e}$ variables.

\paragraph{Nodes mapping constraints}
The constraint \eqref{eq:sumx1} ensures that each vertex of $G_{1}$ is mapped at most to one vertex of $G_{2}$, while the constraint \eqref{eq:sumx2} ensures that each vertex of $G_{2}$ is at most mapped to one vertex of $G_{1}$ :
\begin{subequations}
  \begin{align}
  \sum_{u_k \in V_2} \textbf{v}_{i,k}  \leq 1 \quad \forall u_i \in V_1 \label{eq:sumx1}\\
  \sum_{u_i \in V_1} \textbf{v}_{i,k}  \leq 1 \quad \forall u_k \in V_2 \label{eq:sumx2}
  \end{align}
\end{subequations}

\paragraph{Edges mapping constraints}
Similarly to the vertex mapping constraints, the constraints \eqref{eq:sumy1} and \eqref{eq:sumy2} guarantee a valid mapping between the edges:
\begin{subequations}
  \begin{align}
    \sum_{z_{kl} \in E_2} \textbf{e}_{ij,kl}  \leq 1 \quad \forall z_{ij} \in E_1 \label{eq:sumy1}\\
  \sum_{z_{ij} \in E_1} \textbf{e}_{ij,kl}  \leq 1 \quad \forall z_{kl} \in E_2 \label{eq:sumy2}
  \end{align}
\end{subequations}

\paragraph{Topological constraints}
The respect of the graph topology in the mapping of the vertices and of the edges is described as follows:
\textit{An edge $z_{ij} \in E_1$ can be mapped to an edge $z_{kl} \in E_2$ only if the head vertices $u_i \in V_1$ and $u_k \in V_2$, on the one hand, and if the tail vertices $u_j \in V_1$ and $u_l \in V_2$, on the other hand, are respectively mapped.}
These constraints can be expressed with the following constraints \eqref{eq:topology_3} and \eqref{eq:topology_4}:

\begin{itemize}
 \item Given an edge $z_{ij} \in E_1$ and a vertex $u_k \in V_2$, there is at most one edge whose initial vertex is $u_k$
       that can be mapped with $z_{ij}$:
       \begin{equation}
	  \sum_{z_{kl} \in E_2} \mathbf{e}_{ij,kl} \leq \mathbf{v}_{i,k} \quad \forall u_k \in V_2, \forall z_{ij} \in E_1\\
	  \label{eq:topology_3}
       \end{equation}
 \item Given an edge $z_{ij} \in E_1$ and a vertex $u_l \in V_2$, there is at most one edge whose terminal vertex is $u_l$
       that can be mapped with $z_{ij}$:
       \begin{equation}
	  \sum_{z_{kl} \in E_2} \mathbf{e}_{ij,kl} \leq \mathbf{v}_{j,l} \quad \forall u_l \in V_2, \forall z_{ij} \in E_1\\
	  \label{eq:topology_4}
       \end{equation}
\end{itemize}

\paragraph{The formulation}
The entire formulation is called $GMS$ and described as follows : 
\begin{model}{$GMS$}
\begin{subequations}
    \begin{align}
        &\max_{{\mathbf{v},\mathbf{e}}} SL(\mathbf{v},\mathbf{e}) \label{f2:o}\\
    \text{subject to}\quad
    &\sum_{u_k \in V_2} \textbf{v}_{i,k}  \leq 1 \quad \forall u_i \in V_1 \\
    &\sum_{u_i \in V_1} \textbf{v}_{i,k}  \leq 1 \quad \forall u_k \in V_2 \\
    &\sum_{z_{kl} \in E_2} \textbf{e}_{ij,kl}  \leq 1 \quad \forall z_{ij} \in E_1\\
    &\sum_{z_{ij} \in E_1} \textbf{e}_{ij,kl}  \leq 1 \quad \forall z_{kl} \in E_2 \\
      &\sum_{z_{kl} \in E_2} \mathbf{e}_{ij,kl} \leq \mathbf{v}_{i,k} \quad \forall u_k \in V_2, \forall z_{ij} \in E_1\\
      &\sum_{z_{kl} \in E_2} \mathbf{e}_{ij,kl} \leq \mathbf{v}_{j,l} \quad \forall u_l \in V_2, \forall z_{ij} \in E_1\\
    \text{with}\quad
    &\mathbf{v}_{i,k} \in \{0, 1\} \quad \forall (u_i, u_k) \in V_1 \times V_2\\
    &\mathbf{e}_{ij,kl} \in \{0, 1\} \quad \forall (z_{ij}, z_{kl}) \in E_1 \times E_2
  \end{align}
\end{subequations}
\label{model:F2}
\end{model}

\paragraph{Topology-relaxed formulation}
From Model \ref{model:F2}, a topology-relaxed formulation is obtained by removing the topological constraints defined in the equations \eqref{eq:topology_3} and \eqref{eq:topology_4}. Such a model does not model the original graph matching problem (Problem \ref{prob:GM}) anymore. For a given graph $G$, the vertex set and the edge set are two independent sets. Vertices and edges can be matched independently to each other. There are no constraints to link the matching of vertices with the matching of edges. 

The entire topology-relaxed formulation is called $GMS^*$ and described as follows : 
\begin{model}{$GMS^*$}
\begin{subequations}
    \begin{align}
        &\max_{{\mathbf{v},\mathbf{e}}} SL(\mathbf{v},\mathbf{e}) \\
    \text{subject to}\quad
    &\sum_{u_k \in V_2} \textbf{v}_{i,k}  \leq 1 \quad \forall u_i \in V_1 \\
    &\sum_{u_i \in V_1} \textbf{v}_{i,k}  \leq 1 \quad \forall u_k \in V_2 \\
    &\sum_{z_{kl} \in E_2} \textbf{e}_{ij,kl}  \leq 1 \quad \forall z_{ij} \in E_1\\
    &\sum_{z_{ij} \in E_1} \textbf{e}_{ij,kl}  \leq 1 \quad \forall z_{kl} \in E_2 \\
    \text{with}\quad
    &\mathbf{v}_{i,k} \in \{0, 1\} \quad \forall (u_i, u_k) \in V_1 \times V_2\\
    &\mathbf{e}_{ij,kl} \in \{0, 1\} \quad \forall (z_{ij}, z_{kl}) \in E_1 \times E_2
  \end{align}
\end{subequations}
\label{model:F2degraded}
\end{model}

The Model \ref{model:F2degraded} can be seen as two independent linear sum assignment problems operating on the vertex sets and edges sets, respectively. This relaxed model will help us to better understand the role of topological constraints on the deep model performances.




\subsection{Quality-aware heuristics with mixed-integer linear program solvers}

In computer science and operations research, approximation algorithms are effective algorithms that find approximate solutions to optimization problems (in particular $\mathcal{NP}$-hard problems) with provable guarantees on the distance of the returned solution to the optimal one. This distinguishes them from heuristics, which find reasonably good solutions on some inputs, but provide no clear indication at the outset on when they may succeed or fail.


In general, MILP formulations are solved by black-box solvers such as CPLEX \cite{cplex2009v12}, Gurobi \cite{gurobi}, etc. These solvers are equipped with an arsenal of effective algorithms. Theses algorithms can be applied at two different moments: 1) a preprocessing stage before searching for a solution and 2) during the solving.
For instance, let us mention two preprocessings that are embedded into the solvers. 1) Automatic cuts are sets of constraints added to the model for a given instance. Cuts are expected to reduce the search space (Gomory cuts, Disjunctive cuts, ... ). 2) Variable fixing consists in analyzing an instance and figures out if some variables can be fixed to 0 or 1 in the graph matching model (with respect to Driebek penalty for instance). The aim is to reduce the number of variables to be fed to solving methods.

A MILP instance can be solved by a tree-based methods like \textit{branch and bound} methods that takes advantage of the mathematical formulation. During the search, upper bounds of the maximization problem are computed by continuous relaxation and problem decomposition. In a maximization problem, an upper bound (UB) is the objective function value of an infeasible solution ($\tilde{\mathbf{v}},\tilde{\mathbf{e}}$) of the graph matching problem. An upper bound is always higher or equal than the score of the optimal solution.
In addition, lower bounds can be computed by fast local searches such as greedy search or beam search for instances. A lower bound (LB) is the objective function value of
an admissible solution ($\hat{\mathbf{v}},\hat{\mathbf{e}}$) of the graph matching problem. A lower bound is always lower or equal than the objective value of the optimal solution.

The objective value of the optimal solution is bounded by the upper bound and the lower bound: $LB<optimal<UB$.

For a given instance, the gap level measures how close a pair of lower bound and upper bound are:
\begin{equation}
    \label{gap}
    GAP= \frac{|UB-LB|}{|UB|}
\end{equation}
If the gap level is equal to zero then it means that the lower bound is equal to the upper bound. In such a case, the lower bound is the optimal solution of the problem. At the opposite, if the gap level is high then it means either the upper bound is a bad estimation of the optimal matching or either the lower bound is far from the optimal solution.

For such a reason, the $GAP$ \textit{value} is representative to the \textit{quality level} of a lower bound.

Tree-based methods proceed to an implicit enumeration of all possible solutions without explicitly evaluating all of them by means of an ordered tree. It is constructed dynamically at run time by iteratively creating successor tree nodes called children
nodes. A tree node $p$ here corresponds to a partial matching. The root node of the tree is a special tree node where no graph vertices are matched. From this root node, a global upper bound is computed by continuous relaxation of the MILP. At each iteration, lower bound and upper bounds are computed to the best bounds found so far. At then end of the iteration, the new bounds are involved in the computation of the $GAP$ value. 
A stopping criterion based on the gap level can be added to tree-based methods. The goal is to stop the exploration when a satisfactory GAP value called $\alpha$ is reached. The outputted solution is then the best lower bound found so far. The algorithm is listed in Algorithm \ref{algo:gapexplore}.

$\alpha$ controls the quality of the solution returned by the graph matching solver. If $\alpha=0$ then the solution is an optimal solution and the graph matching solver ($GMS$) can be considered as an exact method. If $\alpha > 0$ then the graph matching solver is an heuristic method with guarantees on the distance of the returned solution to the optimal one. The MIP solver will terminate when the gap between the lower and upper objective bound is less than $\alpha$ times the absolute value of the incumbent lower bound.

\begin{algorithm}
\label{algo:gapexplore}
\LinesNumbered 
\KwData{A graph matching instance: $G_1,G_2,\hat{\mathbf{sv}},\hat{\mathbf{se}}$}
\KwData{$\alpha$: A quality level}
\KwResult{$\hat{\textbf{v}},\hat{\textbf{e}}$ : A matching solution}
\tcp{Compute the root node of the search tree.}
p := ROOT($G_1,G_2,\hat{\mathbf{sv}},\hat{\mathbf{se}}$)\\
\tcp{Compute the upper bound from the root node}
($\tilde{\mathbf{v}},\tilde{\mathbf{e}}$):=ComputeUpperBound(p)\\
UB:=$SL(\tilde{\mathbf{v}},\tilde{\mathbf{e}},\hat{\mathbf{sv}},\hat{\mathbf{se}})$\\
p.UB:=UB\\
\tcp{Compute a lower bound from the root node}
($\hat{\mathbf{v}},\hat{\mathbf{e}}$):=ComputeLowerBound(p)\\
LB:=$SL(\hat{\mathbf{v}},\hat{\mathbf{e}},\hat{\mathbf{sv}},\hat{\mathbf{se}})$\\
\tcp{Compute the GAP value}
p.LB:=LB\\
$GAP:=\frac{|UB-LB|}{|UB|}$\\
\tcp{Add p to the list of tree nodes to be expanded}
OPENLIST:=\{p\}\\
 \While{GAP $>\alpha$}{
    \tcp{Generate all the children nodes from p}
     children := ExpandTreeNode(p)\\ 
     \tcc{Compute an upper bound for each child ($\forall p \in children \quad | \quad Compute \quad p.UB$)}
     ComputeAllUpperBounds(children)\\
     \tcp{Remove $p$ from the list}
     OPENLIST.remove(p)\\
     \tcp{Add the children to the list of tree nodes to be expanded}
     OPENLIST.add(children)\\
     \tcp{Sort in descending order according to the upper bound value}
     OPENLIST.sort()\\
     \tcp{Take the first tree node from the list}
      p := OPENLIST.pop()\\
    \tcp{Compute a lower bound from the current tree node p}
    (${\mathbf{v}},{\mathbf{e}}$):=ComputeLowerBound(p)\\
    \tcp{Check if this lower bound is the best known solution}
    \If{LB $<SL({\mathbf{v}},{\mathbf{e}},\hat{\mathbf{sv}},\hat{\mathbf{se}})$}{
        \tcp{The new solution is better and becomes the new lower bound}
        ($\hat{\mathbf{v}},\hat{\mathbf{e}}$):=(${\mathbf{v}},{\mathbf{e}}$)\\
        LB:=$SL(\hat{\mathbf{v}},\hat{\mathbf{e}},\hat{\mathbf{sv}},\hat{\mathbf{se}})$\\
    }
    
    \tcp{Check if the new upper bound is the best}
    \If{UB $>p.UB$}{
        \tcp{The new upper bound is better and becomes the best upper bound}
        UB:=$p.UB$\\
    }
    

    \tcp{Compute the GAP value}
    $GAP:=\frac{|UB-LB|}{|UB|}$
 }
\Return $\hat{\textbf{v}},\hat{\textbf{e}}$ 
 \caption{Algorithm of the quality-aware heuristic}
\end{algorithm}

\subsection{Differentiation of the graph matching solver}
In this section, we first define the loss function used to compare two assignments. Second, the gradient of the loss is defined. Finally, the gradient of the graph matching solver is defined thanks to a continuous approximation of the loss function.
\paragraph{Loss : Hamming loss function}
In \cite{BB-GM}, the Hamming distance is proposed as loss function between a predicted matching and the ground truth matching of vertices.

\begin{equation}
L( \textbf{v}^{gt}, \textbf{v})= \frac{1}{|V_1|.|V_2| } \Vert \textbf{v}- \textbf{v}^{gt} \Vert_1
\end{equation}
It is more efficient to implement it that way : 
\begin{equation}
L( \textbf{v}^{gt}, \textbf{v})=\frac{1}{|V_1|.|V_2| } \mathbf{1}_{|V_1|.|V_2|}^{T} . (\textbf{v}^{gt}\odot(1-\textbf{v})+(1-\textbf{v}^{gt})\odot\textbf{v})
\end{equation}
$\odot$ is an element-wise product of matrices is known as the Hadamard product. $\mathbf{1}_{|V_1|.|V_2|}$ is vector filled with ones of dimension $|V_1|.|V_2|$.

\paragraph{Gradient of the loss}
The gradient of the loss $L$ with respect to the parameters $\theta$ is defined in Equation \eqref{eq:gradlossgeneral} and is equal to :
\begin{equation*}
\frac{\partial L(\theta)}{\partial \theta} =  \frac{\partial L(\mathbf{ {v} })}{\partial \mathbf{ {v} } } . \frac{\partial GMS( \mathbf{sv}, \mathbf{se})}{\partial  [\mathbf{sv}, \mathbf{se}] } . \frac{\partial NN(\theta)}{\partial \theta }
\end{equation*}
The loss function only depends on node matching variable $\textbf{v}$. So the gradient of the loss will only be computed with respect to $\mathbf{{v}}$. We assume that there is a derivative for the Hamming loss function. We assume that $\frac{\partial L(\mathbf{{v}})}{\partial \mathbf{{v}}}$ is computable.

\paragraph{Gradient of the graph matching solver}

The informative gradient problem (Problem \ref{prob:gradprob}) arises because $L$ is a piecewise constant function. If this piecewise constant function was arbitrary, we would be forced to use zero-order gradient estimation techniques such as computing finite differences. These require prohibitively many function evaluations.
However, the function $L$ is a result of a maximization process induced by $GMS$ and it is known that for smooth spaces $\mathbb{R}^{|V_1|.|V_2|}$ there are techniques for such “differentiation through argmax”. Let us notice that $\mathbf{v} \in \{0,1\}^{|V_1|.|V_2|}$  is a discrete point in a smooth space $\mathbb{R}^{|V_1|.|V_2|}$. So, it turns out to be possible to build – with different mathematical tools – a viable discrete analogy of the continuous case. In particular, from \cite{DBLP:journals/corr/abs-1912-02175}, we can efficiently construct a function $L_\lambda$, a continuous interpolation of $L$.

For the sake of simplicity, we make notations lighter. The loss $L$ depends on $\mathbf{v}^{gt}$ but $\mathbf{v}^{gt}$ does not impact the gradient so we can omit it. By composition, $L$ depends on variables  $\textbf{v},\textbf{e},\textbf{sv},\textbf{se}$, we make these dependencies visible as follows : 
$L(\mathbf{v}^{gt},GMS(\textbf{v},\textbf{e},\textbf{sv},\textbf{se}))=L(\textbf{v},\textbf{e},\textbf{sv},\textbf{se})$

Given by \cite{DBLP:journals/corr/abs-1912-02175} :
\begin{equation}
L_{\lambda}(\textbf{v},\textbf{e},\textbf{sv},\textbf{se}) =  L(\mathbf{\hat{v}}_\lambda)- \frac{1}{\lambda} [SL(\textbf{v},\textbf{e},\textbf{sv},\textbf{se}) - SL(\hat{\textbf{v}}_\lambda,\hat{\textbf{e}}_\lambda,\textbf{sv},\textbf{se}) ]
\end{equation}

Where $\hat{\textbf{v}}_\lambda$ and $\hat{\textbf{e}}_\lambda$ are the solutions of the perturbed/modified combinatorial solver named $GMS_\lambda$. The perturbed graph matching solver is the same as $GMS$ except that the objective function is replaced by $SL_\lambda$.
$$ SL_\lambda(\textbf{v},\textbf{e},\mathbf{sv},\mathbf{se}) = \mathbf{sv}^{T}.\mathbf{v} +  \mathbf{se}^{T}.\mathbf{e} + \lambda L(\mathbf{v})$$
The hyper-parameter $\lambda >0$  controls the trade-off between ''informativeness" of the gradient” and ''faithfulness" to the original function $L$.

\paragraph{Gradient of $L_{\lambda}(\textbf{v}, \textbf{e},\mathbf{sv},\mathbf{se})$}
Now, the gradient of $L_{\lambda}(\textbf{v}, \textbf{e},\mathbf{sv},\mathbf{se})$ can be written as follows :

\begin{equation}
\frac{\partial L_\lambda(\textbf{v}, \textbf{e},\mathbf{sv},\mathbf{se})} {\partial [\mathbf{sv},\mathbf{se}]}= - \frac{1}{\lambda} [\frac{\partial SL(\textbf{v}, \textbf{e},\mathbf{sv},\mathbf{se})}{\partial [ \mathbf{sv},\mathbf{se}]}  -  \frac{\partial SL(\hat{\textbf{v}}_\lambda, \hat{\textbf{e}}L_\lambda,\mathbf{sv},\mathbf{se}) }{\partial [ \mathbf{sv},\mathbf{se} ]}  ]
\end{equation}
According to the equation \eqref{eq:matchingfunctionlinear} where $SL$ is linear in function of $\mathbf{sv}$ and $\mathbf{se}$, we have the following derivative:
\begin{equation}
\frac{\partial L_\lambda(\textbf{v}, \textbf{e},\mathbf{sv},\mathbf{se})} {\partial [ \mathbf{sv},\mathbf{se} ]}= - \frac{1}{\lambda} [\textbf{v}-\hat{\textbf{v}}_\lambda, \textbf{e}-\hat{\textbf{e}}_\lambda ]
\end{equation}

This is in fact the exact gradient of a piecewise linear interpolation of $L$ in which a hyperparameter $\lambda > 0$ controls the interpolation range.

\paragraph{Efficient computation of $GMS_\lambda$}
 In \cite{DBLP:journals/corr/abs-1912-02175}, the authors showed that $GMS_\lambda$ can be computed in a very efficient way without changing the objective function $SL$ of $GMS$ for $SL_\lambda$. Instead of perturbing the objective function, the similarities $\textbf{sv}$, $\textbf{se}$ can be perturbed.

Let $\hat{\textbf{sv}}$, $\hat{\textbf{se}}$ and $\hat{\textbf{v}}$ be fixed and already computed by the \textit{forward pass} of the neural network.

Let us write $\frac{\partial L(\textbf{v})}{\partial \textbf{v}}(\hat{\textbf{v}})$ that means the derivative of $L$ with respect to $\textbf{v}$ at the point  $\textbf{v}=\hat{\textbf{v}}$.

Let $\textbf{sv}_\lambda=\hat{\textbf{sv}}+\lambda \frac{\partial L(\textbf{v})}{\partial \textbf{v}}(\hat{\textbf{v}})$

With the Hamming loss function, $L$ depends only on $\textbf{v}$ and does not depend on $\textbf{e}$. So the derivative of $L$ with respect to $\textbf{e}$ equal 0.

Let $\textbf{se}_\lambda=\hat{\textbf{se}}+\lambda \frac{\partial L(\textbf{v})}{\partial \textbf{e}}(\hat{\textbf{v}})=\hat{\textbf{se}}$

Therefore $\hat{\textbf{v}}_\lambda$,$\hat{\textbf{e}}_\lambda$ :
\begin{equation}
\hat{\textbf{v}}_{\lambda},\hat{\textbf{e}}_{\lambda} = arg \max_{(\textbf{v}_{\lambda},\textbf{e}_{\lambda}) \in \Gamma} SL(\textbf{sv}_\lambda,\textbf{se},\textbf{v}_{\lambda},\textbf{e}_{\lambda}) 
\end{equation}

\subsection{Forward and Backward algorithms}

The forward algorithm propagates the input data in the forward direction through the neural network.
The backward algorithm computes the gradient of the loss function with respect to the parameters of the neural network for a single input–output.

We present the forward and backward algorithms but only for the graph matching layer of the deep graph matching architecture. The other layers remain the same as usual in any neural network.

The forward and backward algorithms are listed in Algorithm \ref{algo:forward} and Algorithm \ref{algo:backward}.

\begin{algorithm}
\label{algo:forward}
\LinesNumbered 
\KwData{$\hat{\mathbf{sv}}$ and $\hat{\mathbf{se}}$ : node-to-node and edge-to-edge similarities. Features provided by the neural network ($NN$)}
\KwResult{$\hat{\textbf{v}},\hat{\textbf{e}}$ : A matching solution}
\tcp{Run the graph matching solver}
$\hat{\textbf{v}},\hat{\textbf{e}} :=  GMS(\hat{\mathbf{sv}},\hat{\mathbf{se}}) $

\tcp{Save matching variables and similarities as needed for backward algorithm}

SAVE $\hat{\textbf{v}},\hat{\textbf{e}}, \hat{\mathbf{sv}}$ and $\hat{\mathbf{se}}$

\Return $\hat{\textbf{v}},\hat{\textbf{e}}$
 \caption{Forward algorithm of the graph matching layer.}
\end{algorithm}

\begin{algorithm}
\label{algo:backward}
\LinesNumbered 
\KwData{$\frac{\partial L(\mathbf{{v}})}{\partial \mathbf{{v}}}(\hat{\mathbf{{v}}})$ : Gradient of the loss with respect to $\mathbf{{v}}$ at the point $\mathbf{{v}}=\mathbf{\hat{v}}$.}
\KwData{$\lambda$ : Parameter of fidelity to the original loss $L$.}
\KwResult{$\frac{\partial GMS( \mathbf{sv}, \mathbf{se})}{\partial  [\mathbf{sv}, \mathbf{se}]}$ : Gradient of the graph matching layer}

\tcp{Load the matching variables and similarities}

LOAD $\hat{\textbf{v}},\hat{\textbf{e}}, \hat{\mathbf{sv}}$ and $\hat{\mathbf{se}}$

\tcp{Calculate the modified similarities}
$\textbf{sv}_\lambda=\hat{\textbf{sv}}+\lambda \frac{\partial L(\textbf{v})}{\partial \textbf{v}}(\hat{\textbf{v}})$

\tcp{Run the graph matching solver}
$\hat{\textbf{v}}_{\lambda},\hat{\textbf{e}}_{\lambda} :=  GMS(\textbf{sv}_\lambda,\hat{\mathbf{se}}) $

\Return $- \frac{1}{\lambda} [\hat{\textbf{v}}-\hat{\textbf{v}}_\lambda ]$
 \caption{Backward algorithm of the graph matching layer.}
\end{algorithm}

The backward algorithm is simple to implement and only runs the solver once on modified input.

\section{Experiments}
In Section 1, we raised several questions about the deep graph matching methods. In response to these problems, we have done several experiments to give practical evidence. 
To avoid the choice of the feature extraction module, we followed the choice of the best known method named BB-GM \cite{BB-GM} and so we adopted the SplineCNN\cite{fey2018splinecnn} to extract the similarity values between nodes and edges. We focuses on the graph matching layer and we put to the test different graph matching solvers. 

This section is split into three parts. First, we present the experimental setting that is based on the experimental setting defined in \cite{BB-GM}. Second, we present the experiments that were carried out and why they can be helpful to answer our questions. Finally, we analyse the results of our experiments to answer the questions about the impact of matching quality on the deep graph matching effectiveness.

\subsection{Experimental setting}
In this subsection, we describe our experimental setting with data sets, compared methods, evaluation metrics, hyper-parameters and the learning protocol. 
\paragraph{Datasets}
Pascal VOC\cite{everingham2010pascal} is a well-known benchmark data set in the Visual Object Classes Challenge which have 20 classes. It contains 7,020 training images and 1,682 testing images in total. Following the practice of BB-GM\cite{BB-GM}, we choose 16,000 samples to train and 1,000 sample of each classes to test. The object in an image is cropped around its bounding box and resize to $256\times 256$ pixels. Each object has several keypoints with Berkeley annotations\cite{bourdev2009poselets}.

SPair-71K\cite{min2019spair}, is a new large-scale benchmark data set of semantically paired images and has 70,958 image pairs. Compared with Pascal 3D+ and Pascal VOC 2012, the source datasets of SPair-71K\cite{min2019spair}, it has richer and comprehensive annotation information, clearer data set splits and higher quality images. We choose 3,200 samples to train and 1,000 sample of each classes to evaluate.

\paragraph{Compared methods}

\begin{table}[]
    \centering
    \begin{tabular}{|p{2.5cm}||p{3.5cm}|p{2.5cm}|} 
    \hline
        Methods & Matching problem & Solver \\ \hline \hline
         DIP-$GM$ (our proposal)&  Graph matching & Combinatorial Exact\\ \hline
         DIP-$GM^*$ (our proposal)&  Two LSAP applied to vertex and edge sets& Combinatorial Exact \\ \hline
        BB-$GM$ \cite{BB-GM}&  Graph matching & Combinatorial Heuristic\\ \hline 
         Sinkhorn Net \cite{SinkhornNet}&  Relaxed LSAP applied to vertex sets& Continuous  Heuristic\\ \hline
    \end{tabular}
    \caption{Compared methods with different matching solvers.}
    \label{tab:comparedmethods}
\end{table}
In our experiments, DIP-GM is the proposal based on our MILP model (Model \ref{model:F2}) of the graph matching problem. It stands for Deep Integer Program-Graph Matching. DIP-GM$^*$ is our proposal based on our topology-relaxed MILP model (Model \ref{model:F2degraded}) where topological constraints are removed. The Model \ref{model:F2degraded} can be seen as two independent linear sum assignment problems operating on the vertex sets and edges sets, independently. The BB-GM \cite{BB-GM} method is also compared in the exact same setting meaning with the same feature extraction module and loss function. It is equipped with the effective heuristic based Lagrangean decomposition \cite{DBLP:journals/corr/SwobodaRAKS16}. All the aforementioned methods operate in the discrete domain and call a combinatorial solver.
At the opposite, the Sinkhorn Net \cite{SinkhornNet}, a famous graph matching algorithm in the deep graph matching field, is also the object of our attention. It reduces the graph matching problem to a relaxed linear sum assignment problem (Problem \ref{prob:relaxedlsap}) applied on two sets of vertices. This problem is solved in the continuous domain by the Sinkhorn algorithm \cite{SinkhornAlgo}. 

All the aforementioned methods are compared in the exact same setting. Except for the Sinkhorn network \cite{SinkhornNet} where the loss is slightly modified because the Hamming loss is not suitable for a continuous solution. For continuous solutions, the permutation loss defined in \cite{PCA-GM} is well adapted and was selected. These methods are reported in Table \ref{tab:comparedmethods}.

For a fair comparison, we also competed with the state of the art methods that differ in terms of losses and feature extraction modules: GMN-PL\cite{GMN} using permutation loss, PCA-GM\cite{PCA-GM}, GLMNet\cite{GLMNet}, $CIE_1$-H\cite{CIE}, DGMC*\cite{DGMC}. Among them, DGMC\cite{DGMC} uses the keypoint intersection sampling strategy and renamed it as DGMC$^*$.

\paragraph{Evaluation measure}
In this paper, we discuss about a matching problem between two graphs. The matching procedure is learned in a supervised manner (Problem \ref{prob:gmlearning}). The ground-truth matching is given by a binary vector ($\mathbf{v}^{gt}$) and the predicted matching is given by ($\mathbf{v}$). The similarity between $\mathbf{v}^{gt}$ and $\mathbf{v}$ is measured by an accuracy score as follows : 
\begin{align}
    \label{7}
    \text{Acc} = \sum (\mathbf{v}_{i,j} * \mathbf{v}^{gt}_{i,j}) / \sum \mathbf{v}^{gt}_{i,j}
\end{align}
$Acc$ is a ratio bounded between zero and one, the larger the better.

\paragraph{Hyper-parameters}
We follow the parameter settings provided by \cite{BB-GM}, in order to accurately reflect the influence of the solver on the model. The optimizer in our architecture is Adam \cite{da2014method}, the initial learning rate is $2\times 10^{-3}$, and it is halved with a fixed step size as the training progresses. Learning rate for the VGG16 weights is the model learning rate multiplied with $10^{-2}$. The batch size of model input is 8. The number of epoch is 10 in order to keep consistent as \cite{BB-GM}. 

The hyper-parameter $\lambda$ is fixed to 80.0 as in \cite{BB-GM}. The MILP solver parameters are set to default values of Gurobi users' manual \cite{gurobi}.

\paragraph{Learning protocol}
The parameters of the deep model are trained by an iterative method called Adam. An epoch is a complete pass through the entire training data. At the last epoch, the model and its parameters are saved. This final deep model is evaluated on the test databases. 

As in \cite{BB-GM}, we used a random sampling strategy for the training set in all experiments. It can raise the generalization ability of the model. 
Pascal VOC \cite{everingham2010pascal} and SPair-71K \cite{min2019spair} have 16,000 samples and 3,000 samples to train, respectively. Both our methods and \cite{BB-GM} are restart 5 times and report the mean.

\paragraph{Runtime and code}
Our experimental equipment is a single RTX-2080Ti GPU and a single Intel Xeon Silver 4216 CPU. The computational bottleneck of our architecture is constructing the mathematical model of Gurobi \cite{gurobi} and evaluating the VGG16 backbone. Approximately 5 to 10 samples per second are processed. The code is published in Github: \url{https://github.com/C-puqing/DIP-GM/}.

\subsection{Experiments}
In this section, experiments are described. We explain why and how we conducted them.

\paragraph{Experiment \#1: } In terms of matching accuracy, we want to compare our exact graph matching solver DIP-GM against heuristics methods. Heuristics methods are the state-of-the-art methods and our DIP-GM$^*$ where the topological constraints Eq.\eqref{eq:topology_3} and Eq.\eqref{eq:topology_4} are dropped. DIP-GM and DIP-GM$^*$ are solved by an exact MILP solver but DIP-GM$^*$ is an heuristic because the graph matching problem is relaxed. We first evaluate the model accuracy on Pascal VOC\cite{everingham2010pascal} dataset with all competing approaches above. Then, we compared with \cite{DGMC} and \cite{BB-GM} on the SPair-71K dataset.

\paragraph{Experiment \#2: }  Then, we want to observe the effect of quality level ($\alpha$) on the matching accuracy. The quality-aware heuristic with MILP solver named DIP-GM$^*$ brings convenience for this goal. We used a group different quality levels to test the effect of the quality of the solver on the final result. This group of $\alpha$ values are $\{0, 0.5, 1, 1.5, 2\}$. We tried many values and finally chose 0.5 as the fixed step size. $\alpha=0$ means that optimal solutions are wanted while $\alpha=2$ denotes that low quality solutions can be returned by the MILP solver. A MILP solver with a stopping criterion $\alpha=2$ can be considered as a ''bad" heuristic because outputted solutions can be are far from the optimal solutions. This experiment was only conducted on Pascal VOC, because it is a more common data set in the deep graph matching literature than SPair-71K. 

\paragraph{Experiment \#3: } Last, we want to oppose the continuous matching method Sinkhorn Net \cite{SinkhornNet} to our combinatorial methods. The Sinkhorn solver \cite{SinkhornAlgo} for the Relaxed LSAP (Problem \ref{prob:relaxedlsap}) is opposed to the exact MILP solver. 

\subsection{Results and analysis}
\paragraph{Experiment \#1: }
\begin{table}[htbp]
    \centering
    \includegraphics[scale=0.3]{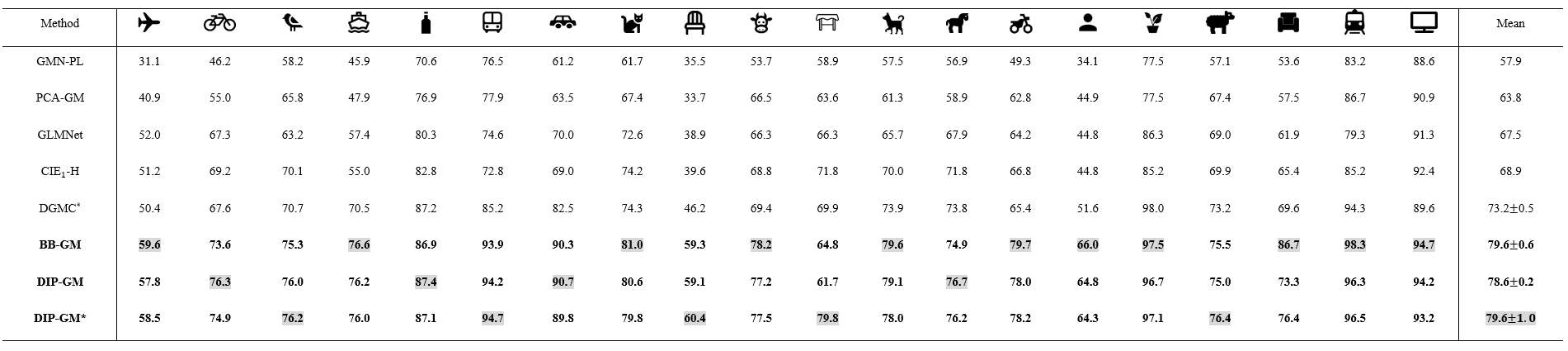}
    \caption{The evaluated matching accuracy(\%) on Pascal VOC using intersection strategy for filtering keypoints. Our methods and BB-GM\cite{BB-GM}, DGMC*\cite{DGMC} are restarted 5 times and report the mean result.}
    \label{t1}
\end{table}

\begin{table}[htbp]
    \centering
    \includegraphics[scale=0.29]{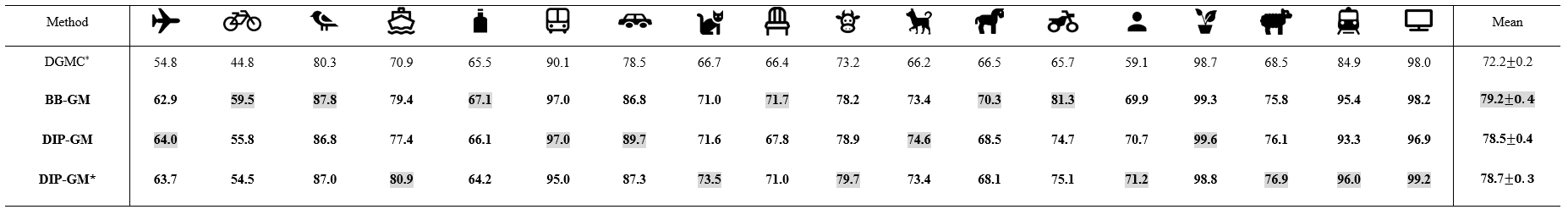}
    \caption{The evaluated matching accuracy(\%) on SPair-71K for all classes. Our methods and BB-GM\cite{BB-GM}, DGMC*\cite{DGMC} are restarted 5 times and report the mean result.}
    \label{t2}
\end{table}

The results about the experiment \#1 are reported in Table \ref{t1} and Table \ref{t2}. It can be seen that BB-GM, DIP-GM and DIP-GM$^*$ achieved the best results among all the state of the art methods. From this statement, we conclude that combinatorial solvers really matter and are important to achieve good results. BB-GM, DIP-GM and DIP-GM$^*$ achieved very similar results and there is no clear winner. The gap in accuracy between the combinatorial heuristic (BB-GM) and the exact solver (DIP-GM) is not decisive. From this result, we say that heuristics can be sufficient to solve the conflicting evidence. However the question about how effective the heuristic should be is answer by the experiment \# 2. The heuristic based on Lagrangean decomposition \cite{DBLP:journals/corr/SwobodaRAKS16} is effective enough to be integrated in such a deep graph matching architecture without degraded the accuracy performance. We also noted that the MILP solver always obtains the optimal solution at the root node of the branch-and-bound search tree (See Algorithm \ref{algo:gapexplore}). From this fact, we conclude that the learned features makes the matching problem simple and so, exact methods are not mandatory. Finally, in average, DIP-GM$^*$ is 1\% better than DIP-GM so it means that relaxing the topological constraints of the graph matching problem have a positive effect on the matching task. We can conclude that the graph topology is not a key feature. Graphs are built from Delaunay triangulation and this structure is not dedicated to the matching task. 

\paragraph{Experiment \#2: }

\begin{figure}[htbp]
    \centering
    \includegraphics[width=\textwidth]{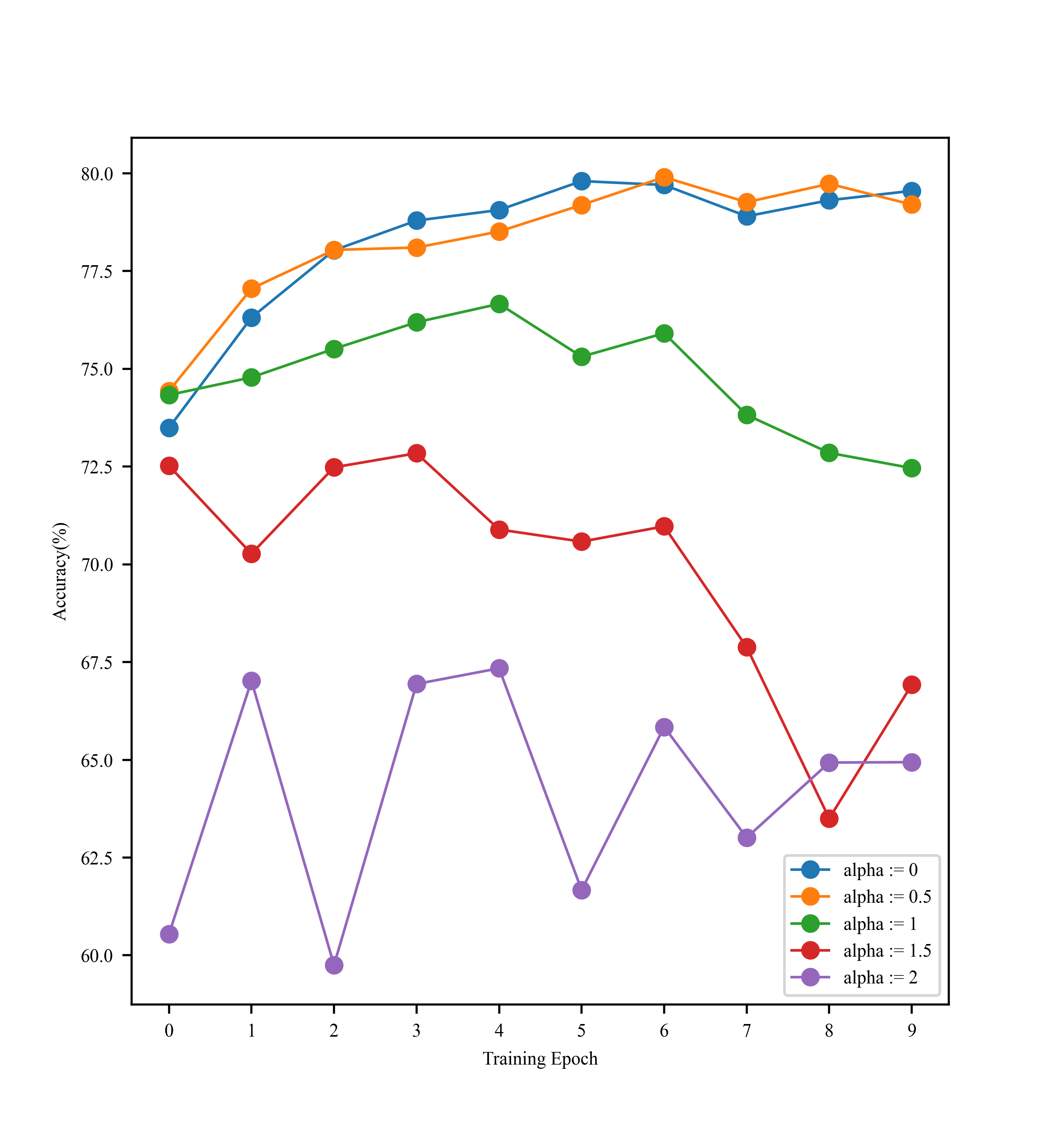}
    \caption{This picture demonstrates the effect of quality level ($\alpha$) on accuracy. The experiment was conducted on Pascal VOC, and the five curves describe the variation of accuracy with the training epoch.}
    \label{fig:2}
\end{figure}

Figure \ref{fig:2} shows that increasing $\alpha$ within an appropriate range will not have a negative impact. However, excessive relaxations seriously affect the performance of the model. This shows that the quality of the solution provided by the graph matching algorithm is very important to the deep model, and it is best not to compromise on difficult combinatorial tasks. 

Especially, at the first epochs, bad quality solutions are very damageable because the feature space is not well fitted to the matching task yet. At the first epochs, the matching problems to be solved are harder than when the parameters ($\theta$) are well adapted to the final task. Starting the training process with bad heuristics can make the training fail.

\begin{figure}[htbp]
    \centering
    \includegraphics[width=\textwidth]{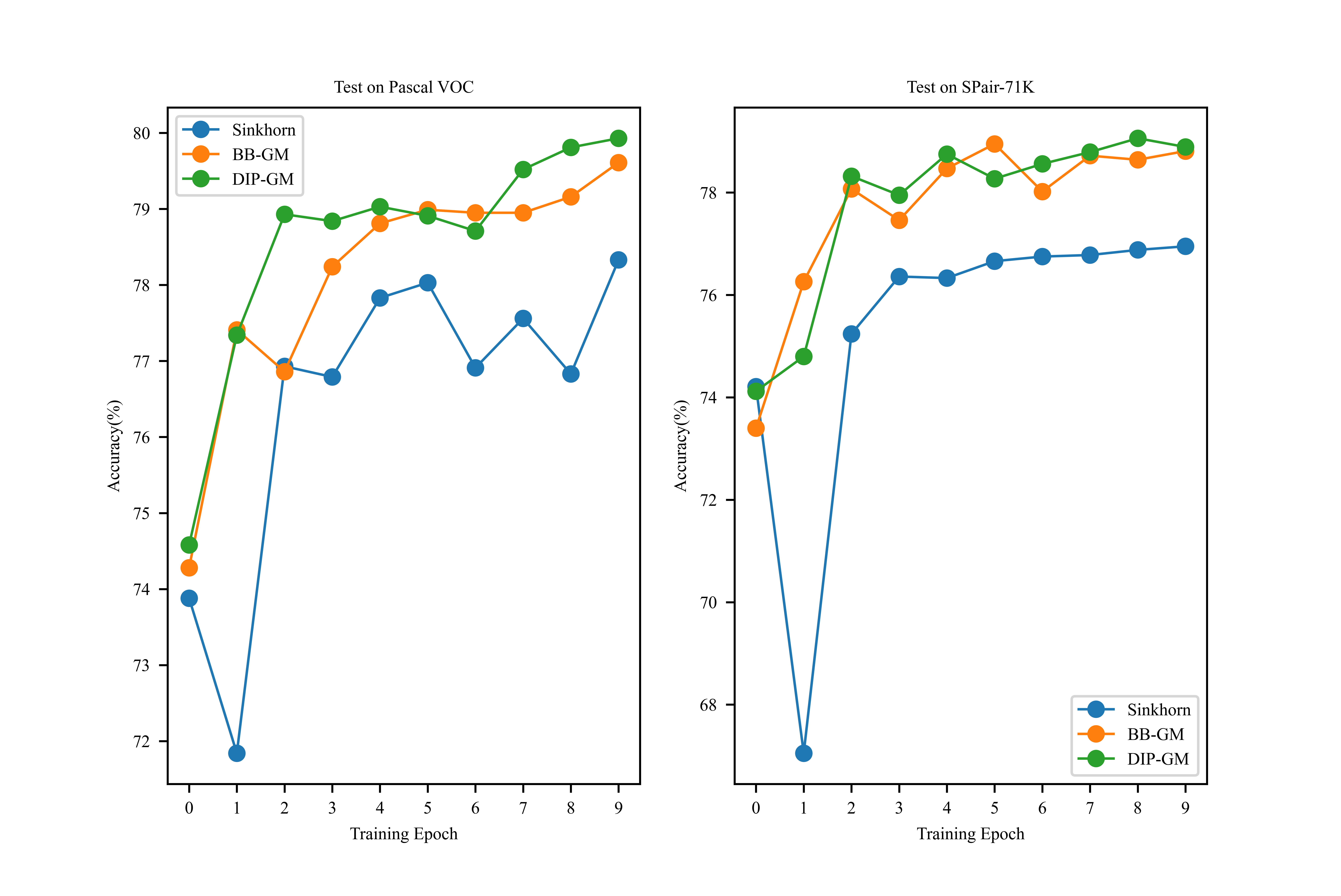}
    \caption{Left shows the result of evaluation on Pascal VOC and right is on SPair-71K.}
    \label{fig:3}
\end{figure}

\paragraph{Experiment \#3: }
Figure \ref{fig:3} reports the results of the experiment. We can conclude that relaxing the graph matching problem to the relaxed LSAP leads to a significant loss of accuracy of 3 \% on average. Discarding edge matching constraints and variables does not mean that structural relationships are not considered at all. Node features extracted by the feature extraction module implies structural information. However, node features cannot compensate the lack of edge features.


\section{Conclusions}
In this paper, an end-to-end deep graph matching model is proposed. The matching problem is decomposed into two parts: extraction of local features and matching at best the local features. Matching at best the local features is modeled as a graph matching problem to consider the relation between local features. Our deep architecture is composed of a combinatorial layer to solve the graph matching problem. An exact mixed integer linear programming solver is integrated in the combinatorial layer. An informative gradient of the combinatorial layer is computed thanks to a continuous interpolation of the loss function. We answer a very important question about how the quality of the solutions provided by the graph matching solver affects the model training. The quality of the solutions depends on the effectiveness of the graph matching solver and the type of relaxations of the original graph matching problem. Our experiments showed that relaxing graph matching problem to some extend does not affect negatively the matching accuracy. It is the case when the graph matching problem is solved by an effective heuristic. Another positive relaxation appears when the graph matching problem is relaxed into two linear assignment problems applied on nodes and edges sets independently. From our experiments, we concluded that learned features reduce the difficulty of solving the matching problem. However, if the graph matching problem is further relaxed to a linear assignment problem applied node sets only then matching accuracy decreases by a margin of 15 \%. In the same way, we characterised as ''bad" heuristics, the heuristics that return solutions with a relative gap between the lower and upper bounds above $\alpha=1$. Experiments showed that with such bad heuristics the matching accuracy decrease of 10 \% in average. That is why deep graph matching should be relaxed carefully. Finally, results showed that combinatorial methods outperformed the state of the art methods operating in the continuous domain by a large margin of 5 \%. These conclusions provides scientific guidance for the design of deep graph matching models.
In future work, other combintorial problems could be investigated to see if our conclusions hold true and if general rules about problem relaxations arise.




\section*{Acknowledgements}
We are grateful to Romain Raveaux's support, contains providing language help, writing assistance, and proofreading the article. Besides, we also thank Zhoubo Xu support experiment equipment. This work is supported in part by the National Natural Science Foundation of China (61762027) and the Natural Science Foundation of Guangxi (2017GXNSFAA198172).

\bibliography{mybibfile}

\section*{Biography}
\textbf{Zhoubo Xu}:  In 2004, she obtained M.Sc. in computer sciences from the Guilin University of Electronic Technology, China. From 2007 to 2011, she had worked on her Ph.D. thesis at Xidian University. Since 2018, she has been a professor in the School of Computer Science and Information Security, Guilin University of Electronic Technology.

\textbf{Puqing Chen}: He is pursuing Master degree at the Guiling University of Electronic Technology. In 2020, He awarded the highest level scholarship from the school.

\textbf{Romain Raveaux}: In 2006, he obtained two M.Sc., one in networking and telecommunication and another one in computer sciences from the University of Rouen (France). From 2006 to 2010, he had worked on his Ph.D. Thesis at the L3I laboratory of the University of La Rochelle. Since 2011, he is an Assistant Professor at the LI lab of the University of Tours. In 2019, he obtained is accreditation to supervise research in Computer Science by defending his contributions on the interplay between combinatorial optimization and machine learning for graph matching and classification.

\textbf{Xin Yang}: She is a postgraduate student at the Guiling University of Electronic Technology, China. She focus on the subgraph matching problem of pattern recognition.

\textbf{Huadong Liu}:  He has obtained his M.Sc. in Control Theory and Control Engineering in 2006 at Guilin University of Electronic Technology, China. He is now a lecture at  at the Guilin University of Electronic Technology, working on graph data and data privacy protection.

\end{document}